\documentclass[letterpaper, 10 pt, conference]{ieeeconf}
\usepackage{times}
\usepackage[pdftex]{graphicx}
\usepackage{amsmath,amssymb,amsopn,amstext,amsfonts}
\usepackage{cancel}
\usepackage[space]{cite}
\usepackage{pdfsync}
\usepackage{balance}
\usepackage{color}
\usepackage{mathtools}
\usepackage{algpseudocode}
\usepackage{algorithm}
\usepackage{bm}

\usepackage{diagbox}
\usepackage{float}
\usepackage{epstopdf}
\usepackage{pifont}
\usepackage{amsmath}
\usepackage{multirow}
\usepackage{url}
\usepackage{verbatim}
\usepackage{booktabs}
\usepackage{graphicx}
\usepackage{subcaption}
\usepackage{threeparttable}  
\usepackage{makecell}
\usepackage[linkcolor=black,citecolor=black,urlcolor=black,colorlinks=true]{hyperref}

\newcommand{\tp}{^{\mathrm{T}}}

\newcommand{\rBrac}[1]{\left({#1}\right)}

\newcommand{\sBrac}[1]{\left[{#1}\right]}

\newcommand{\Norm}[1]{\left\Vert{#1}\right\Vert}

\graphicspath{{}}
\DeclareGraphicsExtensions{.png,.jpg,.eps,.pdf,.PNG}
\IEEEoverridecommandlockouts
\begin{document}

\title{Fast-Racing: An Open-source Strong Baseline for $\mathrm{SE}(3)$ Planning in Autonomous Drone Racing}

\author{{Zhichao~Han}\textsuperscript{1,2,3}, {Zhepei~Wang}\textsuperscript{1,2},{Neng~Pan}\textsuperscript{1,2,3}, {Yi~Lin}\textsuperscript{3}, {Chao~Xu}\textsuperscript{1,2}, and~{Fei~Gao}\textsuperscript{1,2} 
	\thanks{
	\textsuperscript{1} State Key Laboratory of Industrial Control Technology, Institute of Cyber-Systems and Control, Zhejiang University, Hangzhou 310027, China.}
	\thanks{
	\textsuperscript{2} Huzhou Institute of Zhejiang University, Huzhou 313000, China.}
	\thanks{
	\textsuperscript{3} DJI, Shenzhen 510810, China.	}
	\thanks{Zhichao Han and Neng Pan completed the real-world experiment in DJI under the mentorship of  Yi Lin.}
	\thanks{E-mail: {\tt\small \{zhichaohan, wangzhepei, panneng\_zju, cxu, fgaoaa\}@zju.edu.cn, yi.lin@dji.com}}
}
\maketitle

\begin{abstract}
	With the autonomy of aerial robots advances in recent years, autonomous drone racing has drawn increasing attention.
	In a professional pilot competition, a skilled operator always controls the drone to agilely avoid obstacles in aggressive attitudes, for reaching the destination as fast as possible.
	Autonomous flight like elite pilots requires planning in $\mathrm{SE}(3)$, whose non-triviality and complexity hindering a convincing solution in our community by now.
	To bridge this gap, this paper proposes an open-source baseline, which includes a high-performance $\mathrm{SE}(3)$ planner and a challenging simulation platform tailored for drone racing.
	We specify the $\mathrm{SE}(3)$ trajectory generation as a soft-penalty optimization problem, and speed up the solving process utilizing its underlying parallel structure.
	Moreover, to provide a testbed for challenging the  planner, we develop delicate drone racing tracks which mimic real-world set-up and necessities planning in $\mathrm{SE}(3)$.
	Besides, we provide necessary system components such as common map interfaces and a baseline  controller, to make our work plug-in-and-use.
	With our baseline, we hope to future foster the research of $\mathrm{SE}(3)$ planning and the competition of autonomous drone racing.
\end{abstract}

\IEEEpeerreviewmaketitle
\section{Introduction}
\label{sec:introduction}
As the capability of sensing and computing advances, the community is continuously pushing the boundary of autonomy of unmanned aerial vehicles (UAVs), making them fly into many industrial and commercial scenarios.
In recent years, researchers have put attentions to a challenging problem, autonomous drone racing, such as IROS (Intelligent Robots and Systems) Autonomous Drone Race~\cite{MoonChallenges} and AlphaPilot~\cite{foehn2020alphapilot}.
Drone racing~\cite{LI2020103621} is a competition where human operators are required to pilot the drone to avoid obstacles as fast as possible in a given tough track.
\par
Autonomous drone racing interleaves many technologies including dynamic modeling, flight control, state estimation, and planning; and provides a good testbed to trial these modules in extreme situations. As the maturity of UAV design and control, state estimation, and planning have been the bottleneck of drone racing for a long time.
Recently, several visual-inertial odometry (VIO) benchmarks and datasets~\cite{antonini2018blackbird}\cite{DelmericoUZH} for high-speed drone racing have been presented publicly, which significantly boost state estimation for aggressive motion.
\par
\begin{figure}[t]
	\vspace{0.6cm}
	\centering
	\begin{subfigure}{1.0\linewidth}
		\centering
		\includegraphics[width=1.0\linewidth]{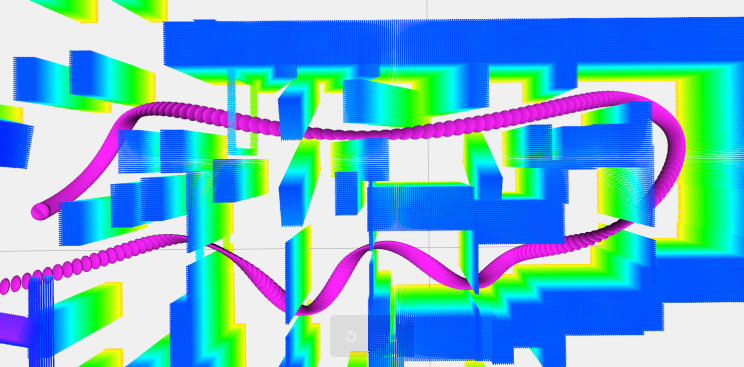}
		\captionsetup{font={small}}
		\caption{The drone flies through narrow gaps and passages. }
		\label{pic:top_graph_a}
		\vspace{0.15cm}
	\end{subfigure}
	\begin{subfigure}{1.0\linewidth}
		\centering
		\includegraphics[width=1.0\linewidth]{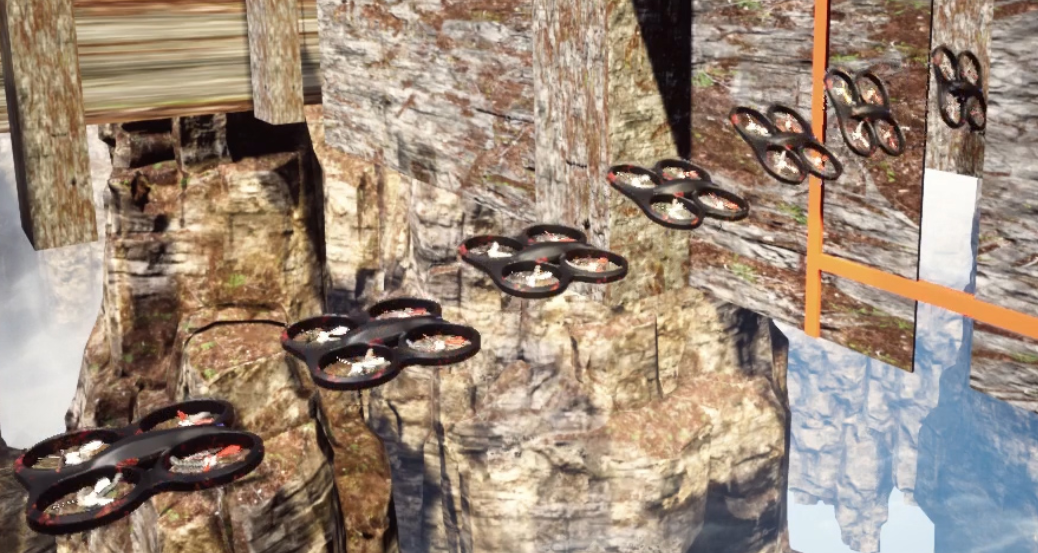}
		\captionsetup{font={small}}
		\caption{Visualization of the $\mathrm{SE}(3)$ motion in our simulator.}
		\label{pic:top_graph_b}
	\end{subfigure}
	\captionsetup{font={small}}
	\caption{
	The two figures show the capability of our $\mathrm{SE}(3)$ planner and one of our environments.
	}
	\vspace{-0.4cm}
	\label{pic:top_graph}
\end{figure}
However, there is always a lack of complete benchmarks for the planning module in drone racing.
Investigating the motion planning problem for drone racing, it is necessary to plan in $\mathrm{SE}(3)$ for fully exploiting the maneuverability of drones to pass challenging tracks with high aggressiveness.
Imagining the drone racing situation, professional pilots always control the drone to aggressively fly against obstacles in a large attitude to reach the destination speedily\footnote{Drone Racing League \url{https://www.youtube.com/watch?v=GTifvVZBNWs}}.
Besides, birds flying at high speeds in a dense forest, such as skylarks, always alternate their attitudes to avoid obstacles.
All evidence shows that $\mathrm{SE}(3)$ planning is imperative, no matter for human piloting or biological flight in nature. Nevertheless, there exists no open-sourced baseline for $\mathrm{SE}(3)$ planning at the moment, which is not conducive to $\mathrm{SE}(3)$ planning research.
The reason is that $\mathrm{SE}(3)$ planning is much more complicated than traditional planning in $\mathbb{R}^3$ because the former requires group operations while conventional optimizers require decision variables in Euclidean space. Moreover, an appropriate simulation environment is necessary to verify the planning algorithm.
However, existing autonomous drone racing simulation environments are not challenging for the planning module, such as tracks in \cite{madaan2020airsim}.
To address the above issues, we propose a strong baseline for $\mathrm{SE}(3)$ planning.
Our $\mathrm{SE}(3)$ planner  generates  safe trajectories with arbitrary $\mathrm{SE}(3)$ constraints, as shown in Fig.\ref{pic:top_graph_a}.
Then, we build static but still highly challenging racing tracks and finish them with our planner. It is worth mentioning that we use a known map and obtain the ground-truth state of the robot from the simulation platform to ensure that the performance of
the planner can be evaluated accurately without disturbance. Similar to Airsim Drone Racing Challenge~\cite{madaan2020airsim}, our planner generates a global trajectory offline before the racing starts while the controller runs online to control the drone to follow the trajectory.
\par
In our paper, we firstly model a quadrotor by its shape and specify the $\mathrm{SE}(3)$ trajectory optimization problem formulation.
Then, following \cite{Wang2021GCOPTER}, we simplify the trajectory optimization without sacrificing optimality 
to solve it cheaply and efficiently.
Moreover, we implement a parallel architecture to speed up the computation.
In addition, we build a highly challenging drone racing simulation platform that contains tracks, map interfaces, and a controller.
Furthermore, we apply our planner in the drone racing tracks as the baseline and show the results for comparisons.
Additionally, we  integrate our planner into an autonomous system where all modules run online, then perform real-world experiments to validate it.
To sum up, contributions of this paper are:
\begin{itemize}
	\item [1)]
	We propose a general $\mathrm{SE}(3)$ collision-free constraint form for 
	multirotor in all shapes.
	\item [2)]
	We implement a high-performance  trajectory optimization based on parallel computing.
	\item [3)]
	We present a plug-in-and-use simulation platform specially designed for $\mathrm{SE}(3)$ planning.
	\item [4)]
	We release the proposed $\mathrm{SE}(3)$ planner and simulation publicly to the community, for fostering future research in $\mathrm{SE}(3)$ planning and autonomous drone racing\footnote{ \url{https://github.com/ZJU-FAST-Lab/Fast-Racing}}.
\end{itemize}
\section{Related Works}
\label{sec:related_works}
\subsection{Autonomous Drone Racing}
Drone racing attracts wide public attention in recent years and also develops significantly with the improvement of algorithms and computing.
The AlphaPilot Challenge\cite{foehn2020alphapilot} is the largest one recently, which builds the qualification round based on the FlightGoggles simulation framework~\cite{MitFlightGooles}. Then, the winners are required to finish the hardware implementation in a later competition.
In addition, IROS conference also hosts an annual hardware drone racing challenge\cite{MoonChallenges} since 2016. The competition consists of navigating autonomously through a number of gates.
In IROS 2018 autonomous drone race, the UZH team showed a winning performance\footnote{IROS 2018 Autonomous Drone Race \url{https://www.youtube.com/watch?v=9AvJ3-n-82w}} whose drone took 30 seconds to complete the track.
The Airsim Drone Racing Lab \cite{madaan2020airsim} introduces a drone racing simulation framework, for bridging the gap between autonomous systems and researchers who specialize in software.
Moreover, it hosts a competition with three tiers, which focus on perception, trajectory planning, and control, respectively.
\par
Drone racing involves lots of aspects such as flight control, state estimation, and motion planning. High-quality and accessible benchmarks are key to fostering the research community forward. In the domain of state estimation, open-sourced datasets (EuRoC\cite{MichaelEuRoc}, MVSEC\cite{Zhu_2018}, Zurich Urban MAV Dataset\cite{ZurichUrban}) boost the development of visual-based algorithms in extreme situations.
In addition, the recently released Blackbird dataset\cite{antonini2018blackbird} and UZH-FPV dataset\cite{DelmericoUZH}, which both target high-speed drone racing, build a foundation for aggressive motion estimation.
\par
\subsection{$\mathrm{SE}(3)$ Planning}
Motion planning is a crucial part of drone racing and
has made excellent progress in recent years. Trajectories of UAVs or other differentially flat systems\cite{MellingerMiniSnap}  are usually parameterized as piecewise polynomials since their derivatives can be used to obtain explicit expressions for the system states and control inputs\cite{MichielFlatSystems}. 
Mellinger et al.\cite{MellingerMiniSnap} present a trajectory smoothness metric, and use anchor waypoints along the trajectory to ensure collision avoidance.
Gao et al.\cite{fei2018icra} construct a flight corridor that consists of a sequence of connected free-space 3-D grids as the safety constraint, and use Bernstein basis polynomial to represent the trajectory.
%\citet{zhou2019robust} apply kinodynamic path searching as the front end and use nonlinear trajectory optimization as the back end based on the ESDF (Euclidean Sign Distance Field).\citet{zhou2020egoplanner} improve \cite{zhou2019robust} by replacing the collision term in penalty function with a new formulation which devised by comparing the colliding trajectory with a collision-free guiding path. Then the ESDF map is not required, which improve the efficiency of the trajectory generation.
Although these works\cite{MellingerMiniSnap}\cite{fei2018icra}\cite{zhou2019robust}\cite{zhou2020egoplanner} generate trajectories efficiently, they require planning in a configuration space where the drone's attitude is neglected.
As described in Sect.~\ref{sec:introduction}, $\mathbb{R}^3$ planners are much inferior when applied to extremely challenging environments or trying to beat elite pilots.
These environments, for instance, contain multiple narrow and tilted gaps that are only traversable if the drone drifts with large attitudes.
To this end, several works~\cite{Falanga,Loianno,Hirata} have been proposed.
However, these works all over-simplify the $\mathrm{SE}(3)$ Planning problem, and only generate trajectories with fixed attitude constraints.
Instead of exploring the whole $\mathrm{SE}(3)$ solution space, they are dominated by the boundary conditions of attitude, thus only applicable to rather simple environments such as only one titled gap exist.
Based on kinodynamic searching, Liu et al.\cite{liu2017searchbased} present a $\mathrm{SE}(3)$ planner which is resolution complete.
However, this method is not applicable to challenging situations in practice.
Since it suffers from a combinatorial explosion, a fine resolution quickly induces unacceptable computing overhead, in both time and memory.
Recently, Wang et al.\cite{Wang2021GCOPTER} propose a planning framework that generates  high-quality SE(3) trajectories even in complicated scenes.
This method shows an efficient way to handle general spatial and temporal constraints in trajectory generation for multirotor.
Still, Wang et al.\cite{Wang2021GCOPTER} do not fully squeeze the potential and finalize the implementation of their method, making it consume unnecessary computation in large-scale problems.
In this paper, we exploit the parallel nature of the optimization program \cite{Wang2021GCOPTER}, and further extend it to incorporate a drone with general shapes.
In this way, we significantly improve the performance and scalability of the $\mathrm{SE}(3)$  trajectory planning, and finally present an easy-to-use baseline.
\section{A High-performance $\mathrm{SE}(3)$ Planner}
\label{sec:se3planning}
In this section, we introduce our $\mathrm{SE}(3)$ trajectory planner with explicit spatial constraints. Our planner inputs are the grid map, the initial and goal states of the robot, while the output is a feasible $\mathrm{SE}(3)$ trajectory. First, we present a general way to model a quadrotor and derive $\mathrm{SE}(3)$ optimization formulation with attitude constraints. Then, we propose a parallel architecture to solve the optimization problem efficiently.
\subsection{Safety Constraint }
\label{subsec:safe_constraint}
In this part, we use a flight corridor to represent the safety constraint which is used in trajectory optimization
to ensure obstacle avoidance. The generation steps are as follows:
\begin{itemize}
	\item [1)]
	A path search is adopted to obtain a feasible path from the start to the goal.
	\item [2)]
	%We start at the first point $w_p$ of the path and follow the path to find the farthest but no more than limited distance unblocked point $w_f$ which indicates the line between them is collision-free.
	We start at the first point $w_p$ of the path and follow the path to find the farthest unblocked point $w_f$ but not exceeding the limit distance. (The unblocked point indicates the line between $w_p$ and $w_f$ is collision-free.
	The limit distance is a specified parameter whose purpose is to restrict the length of the convex hull in an axis.)
	\item [3)]
	RILS \cite{S.liuCorridor} is used to generate a convex polyhedron based on the line in step 2).
	\item [4)]
	The interval point of the line is regarded as  $w_p$ again.
	Then repeating back to step 2) until the endpoint is included in the convex hull.
\end{itemize}
The whole algorithm is shown in Alg.\ref{alg:corridor generation}.\begin{algorithm}[ht]
	\begin{algorithmic}[1]
		\caption{Flight Corridor Generation}
		\label{alg:corridor generation}
		\State Initialize()
		\State $path$ $\leftarrow$ PathSearch($start$,$end$)
		\State $w_p$ $\leftarrow$ $start$
		\While{True}
		\State $w_f$ $\leftarrow$ FindFarPoint($w_p$,$path$,$MaxDis$)
		\State $FlightCorridor$.pushback(ConvexGen($w_p,w_f$))
		\If{$end$ in $FlightCorridor$} 
		\State return $FlightCorridor$
		\EndIf
		\State $w_p \leftarrow$ IntervalPt($w_p$,$w_f$)		
		\EndWhile
	\end{algorithmic}
\end{algorithm}
\subsection{Problem Formulation}
\begin{figure}[t]
	\centering
	\includegraphics[width=0.8\linewidth]{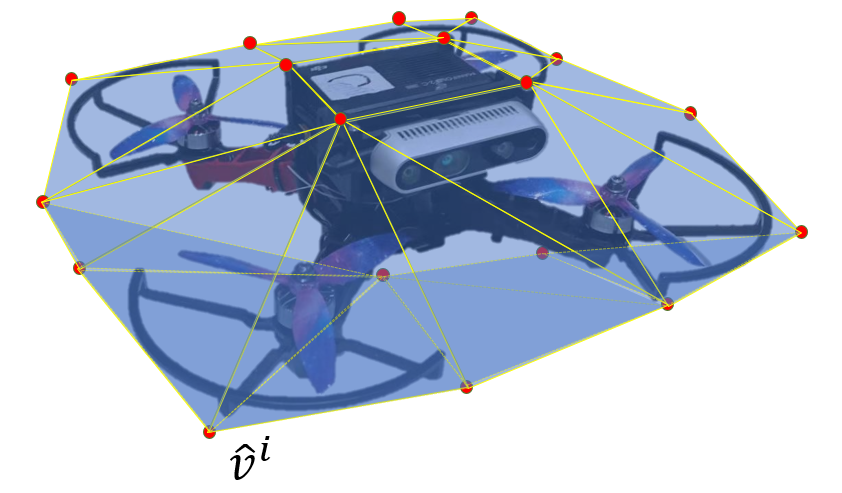}
	\captionsetup{font={small}}
	\caption{
		The drone is completely wrapped in a convex polyhedron. The red points are the vertices of the bounding volume.
	}
	\label{pic:kdop}
	\vspace{-1.0cm}
\end{figure}
\label{sec:Problem Formulation}
Our purpose is to generate a safe $\mathrm{SE}(3)$ trajectory with attitude constraints, in which we use a rotation matrix $R_b$ to represent the quadrotor's attitude. Thanks to the differential flatness property, $R_b$ can be written as an algebraic function
of finite derivatives of its four differentially flat outputs $O_f = [p_x,p_y,p_z,\psi]^{\rm T}:=[\mathbf{x}^{\rm T},\psi]^{\rm T}$ explicitly.
Here $\mathbf{x} = [p_x,p_y,p_z]^{\rm T}$ is the coordinate of the center of mass in the world frame, and $\psi$ is the yaw angle. Utilizing the result from \cite{MellingerMiniSnap}, $R_b$ is as follows:
\begin{small}
\begin{align}
f_b &=  \ddot{\mathbf{x}} + g_we_3, \\
r_{bz} &= f_b/\left\|f_b\right\|_2,  \\
r_{ix} &= \sBrac{\cos{\psi},\sin{\psi},0}\tp,	\\
r_{by} &= \frac{r_{bz} \times r_{ix}}{\left\|r_{bz} \times r_{ix}\right\|_2},\\
r_{bx} &= r_{by} \times r_{bz},\\
R_b &= \sBrac{r_{bx},r_{by},r_{bz}},
\end{align}
\end{small}where $g_w$ is the gravitational acceleration and $e_3=\sBrac{0,0,1}\tp$.
Then, we model the quadrotor as a convex polyhedron which can be generated by using k-DoP (the Boolean intersection of extents along k directions) or specified manually, as shown in Fig.\ref{pic:kdop}.	We define the $i$-th vertex ${\hat v}^i$ of the bounding volume at time $t$ as:
\begin{small}
\begin{equation}
\begin{aligned}
{\hat v}^i(t) = R_b(t)q_{v}^i+\mathbf{x}(t) \ \ 	\forall i \in \left\{1,2,\cdots,K  \right\},
\end{aligned}
\end{equation}
\end{small}where $K$ is the number of vertexs and $q_{v}^i=[{q_{v}^i}_x,{q_{v}^i}_y,{q_{v}^i}_z]^{\rm T}$ is the coordinate of the $i$-th  vertex in the body frame.
Note $q_{v}^i$ is constant once the quadrotor model is identified.

In Sect.~\ref{subsec:safe_constraint}, we build a flight corridor as a series of closed convex polyhedra, defined as:
\begin{small}
\begin{equation}
\begin{aligned}
&\mathcal{P}_j = \left\{
\mathbf{x} \in \mathbb{R}^3 |
(\mathbf{x}-\hat {p}^k_j)\tp\vec{n}^k_j\leq 0,
k = 1,2,\cdots,N_j
\right\}, \\
&\forall j \in \{1,2,\cdots, M\},
\end{aligned}
\end{equation}
\end{small}where $M$ is the number of polyhedra and each polyhedron $\mathcal{P}_j$ consists of $N_j$ hyperplanes which can be represented by a normal vector $\vec{n}^k_j$ outwards and a point $\hat {p}^k_j$ on it.
Constraining the whole body of a quadrotor inscribed by a convex polyhedron written as:
\begin{small}
\begin{equation}
\begin{aligned}
&{(\vec{n}^k_j)}^{{\rm T}}({ {\hat v}^i}(t)-  \hat{p}^k_j) \leq 0, \\
&\forall t \in [T_{j-1},T_j], \forall j \in \left\{1,\cdots,M \right\}, \\ 
&\forall i \in \left\{1,\cdots,K \right\}, \forall k \in \left\{1,\cdots,N_j \right\}. 
\end{aligned}
\end{equation}
\end{small}
\begin{figure*}[t]
	\centering
	\begin{subfigure}{0.2626\linewidth}
		\includegraphics[width=1\linewidth]{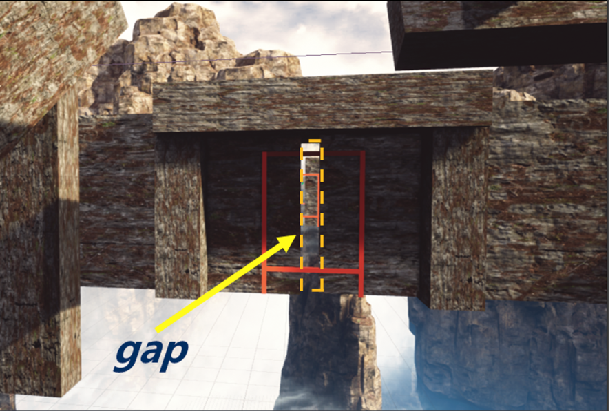}
		\captionsetup{font={small}}
		%\caption{Path searching.}
		\label{pic:left}
	\end{subfigure}
	\begin{subfigure}{0.37548\linewidth}
		\includegraphics[width=1\linewidth]{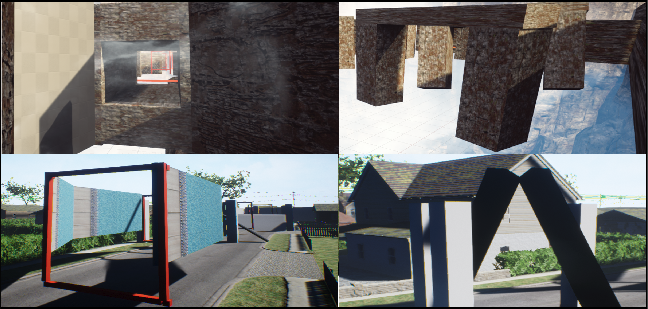}
		\captionsetup{font={small}}
		%\caption{Trajectory generation.}
		\label{pic:middle}
	\end{subfigure}
	\begin{subfigure}{0.2618\linewidth}
		\includegraphics[width=1\linewidth]{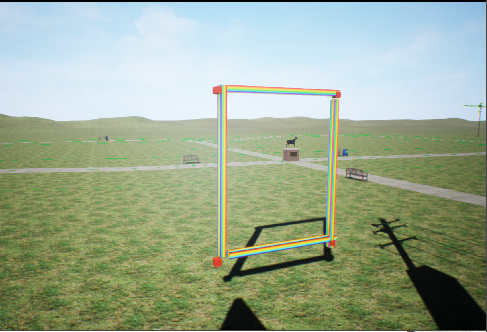}
		\captionsetup{font={small}}
		%\caption{Trajectory generation.}
		\label{pic:right}
	\end{subfigure}
	\captionsetup{font={small}}
	\caption{
		Some features of our drone racing tracks: Left - A gap which is narrower than drone's diameter in track(a), Middle - A few snapshots from our racing environments, Right - The goal gate.
	}
	\label{pic:features}
	\vspace{-0.2cm}
\end{figure*}We adopt piecewise-polynomials to represent our trajectory in each dimension out of ${x, y, z}$:
\begin{small}
\begin{equation}
\mathbf{x}(t)=\left\{
\begin{aligned}
p_1(t)=\sum_{n=0}^{2s-1} &{c_{1n}} {(t-T_0)}^{n}	\ \ 	~~~~T_0 \leq t <T_1,\\
%p_2(t)=\sum_{n=0}^{2s-1} &{c_{2n}} {(t-T_1)}^{n}		\ \	T_1 %\leq t <T_2\\
&\vdots~~~~~~~~~~~~~~~~~~~~~~~~~~~~~\vdots  \\
p_M(t)=\sum_{n=0}^{2s-1} &{c_{jn}} {(t-T_{M-1})}^{n}        \ \  T_{M-1} \leq t <T_{M}.
\end{aligned}
\right.
\end{equation}
\end{small}The desired trajectory is optimized on both the control effort and time regularization to ensure smoothness and aggressiveness.
Moreover, for enforcing safety, spatial constraints are added along the whole trajectory.
Also, feasibility constraints are necessary to guarantee that the trajectory is physically feasible. The optimization problem formulation is:\begin{small}\begin{align}
&\min_{u_j(t), \forall j} \ H = \sum_{j=1}^{M}{\int_{T_{j-1}}^{T_j}{u_j(t)^{\rm T} u_j(t)dt}} + \rho (T_M-T_0) \nonumber\\
&~~~~~~~+\underbrace{ W_v\sum_{j=1}^{M}{\int_{T_{j-1}}^{T_j}{ \mathcal{K}\rBrac{||p_j^{(1)}(t)||_2^2 - v_{max}^2}dt}}}_{\textit{velocity \ limit \ penalty}}  \nonumber  \\
&~~~~~~~+\underbrace{W_a\sum_{j=1}^{M}{\int_{T_{j-1}}^{T_j}{  \mathcal{K}\rBrac{||p_j^{(2)}(t)||_2^2 - a_{max}^2}dt}}}_{\textit{acceleration \ limit \ penalty}}  \nonumber \\
&~~~~~~~+\underbrace{W_c\sum_{j=1}^{M}{\int_{T_{j-1}}^{T_j}{
			\sum_{i=1}^{K}{\sum_{k=1}^{N_j}{
					\mathcal{K}\rBrac{({\hat v}^i(t)-  \widetilde{p}^k_j)\tp\vec{n}^k_j}dt }}}}}_{\textit{collision  \ penalty}},
\label{con:objective function}
\\
&s.t. ~~~\ u_j(t)  = p_j^{(s)}(t),	\ \forall t \in [T_{j-1},T_j],~~~~~~~~~~~~~~
\\
& ~~~~~~~~p_j^{[d]}(T_j) = p_{j+1}^{[d]}(0),~p_j^{[\widetilde{d}]}(T_j) = \overline{p}_j,
\\
& ~~~~~~~~p_j(T_j)   \in {\mathcal{C}_j \cap \mathcal{C}_{j+1}},~T_j-T_{j-1}>0,
\\
& ~~~~~~~~\forall j \in \{1,\cdots,M\},
\\
&~~~~~~~~ p_1^{[s-1]}(0) = \overline{p}_0,\ p_M^{[s-1]}(T_M) = \overline{p}_f        ,\
\end{align}
\end{small}where $\mathcal{K}(x)=\max\rBrac{x,0}^3$ is the cubic penalty, $\rho$ is the weight of time regularization, $W_v, W_a, W_c$ are the corresponding penalty weights, $\overline{p}_j \in \mathbb{R}^{3(\widetilde{d}+1)}$ is the interval condition,
$\overline{p}_0, \overline{p}_f \in \mathbb{R}^{3s}$ are the initial and final states, respectively.
Moreover, $s$ is chosen as $3$, which means that the integration of jerk is used to represent
the smoothness of a trajectory. We define the smoothness cost $\sum_{j=1}^{M}{\int_{T_{j-1}}^{T_j}{u_j(t)^{\rm T} u_j(t)dt}}$ as $\mathcal{S}$ to simplify the formula.
\subsection{Gradient Calculation}
\label{sec:GraCal}
We denote the $j$-th piece of the polynomial trajectory as:
\begin{small}
\begin{equation}
p_j(t) = \mathbf{C}_j^{\rm T}\beta(t-T_{j-1}), \ t \in [T_{j-1},T_j]
\end{equation}
\end{small}where $\mathbf{C}_j \in \mathbb{R}^{2s\times 3 }$ is the coefficient matrix of the piece and	$\beta(t)={[1,t,\cdots,{t}^{2s-1}]}^{\rm T}$ is the time vector. We define the coefficient matrix $\mathbf{C}$ and time allocation vector $\mathbf{T}$:
\begin{small}
\begin{align}
\mathbf{C} &= {(\mathbf{C}^{\rm T}_1,\mathbf{C}^{\rm T}_2,
	\cdots,\mathbf{C}^{\rm T}_M)}^{\rm T} \in \mathbb{R}^{2Ms\times 3},\\
\mathbf{T} &= {(T_1-T_0,T_2-T_1,\cdots,T_M-T_{M-1})}^{\rm T } \nonumber \\
&={(\delta_1,\delta_2,\cdots,\delta_M)}^{\rm T}\in \mathbb{R}^{M}_+.
\end{align}
\end{small}Therefore, the objective function (\ref{con:objective function}) is the function of $\mathbf{C}$ and $\mathbf{T}$: $H = H(\mathbf{C},\mathbf{T})$.
%Also, we denote the velocity, acceleration, jerk and attitude at timestamp $t$ of the $j$-th piece of the trajectory %as $v_j(t)=[{v_j}_x(t),{v_j}_y(t),{v_j}_z(t)]^{\rm T}$, $a_j(t)=[{a_j}_x(t),{a_j}_y(t),{a_j}_z(t)]^{\rm T}$,
%$j_j(t)=[{j_j}_x(t),{j_j}_y(t),{j_j}_z(t)]^{\rm T}$ and 	%$R_{b_j}(t)={[r_{{bx}_j}(t),r_{{by}_j}(t),r_{{bz}_j}(t)]}^{\rm T}$.	
Then we derive the gradient w.r.t. $\mathbf{C}$ and $\mathbf{T}$ as:
\begin{small}
\begin{align}
&\frac{\partial H(\mathbf{C},\mathbf{T})}{\partial \mathbf{C}}\approx
\frac{\partial \mathcal{S}}{\partial \mathbf{C}}+\frac{\partial (E_v+E_a+E_c)}{\partial\mathbf{C}}
\label{equ:H/C},\\
&\frac{\partial H(\mathbf{C},\mathbf{T})}{\partial \mathbf{T}}\approx
\frac{\partial \mathcal{S}}{\partial \mathbf{T}}+\rho\mathbf{1}+\frac{\partial(E_v+E_a+E_c)}{\partial\mathbf{T}},
\label{equ:H/T}\\
&E_v=W_v\sum_{j=1}^{M}{\sum_{l=0}^{L-1}{  \mathcal{K}\rBrac{\Norm{\mathbf{C}_j^{\rm T}\beta^{(1)}(\frac{l}{L}\delta_j)}_2^2 - v_{max}^2}}} \frac{\delta_j}{L} ,\\
&E_a = W_a\sum_{j=1}^{M}{\sum_{l=0}^{L-1}{  \mathcal{K}\rBrac{\Norm{\mathbf{C}_j^{\rm T}\beta^{(2)}(\frac{l}{L}\delta_j)}_2^2 - a_{max}^2}}}\frac{\delta_j}{L} , \\
&E_c=W_c\sum_{j=1}^{M}{\sum_{l=0}^{L-1}{
		\sum_{i=1}^{K}{\sum_{k=1}^{N_j}{
				\mathcal{K}\rBrac{\rBrac{{\hat v}^i_{j}(\frac{l}{L}\delta_j)-  \widetilde{p}^k_j}\tp\vec{n}^k_j}}}}}\frac{\delta_j}{L} ,\\
&{\hat v}^i_{j}(t)={R_b}_j(t)q_{v}^i+\mathbf{C}_j^{\rm T}\beta(t),
\end{align}
\end{small}where $\mathbf{1} \in \mathbb{R}^{M}$ is an all-ones vector, $R_{b_j}(t)$ is the attitude at timestamp $t$ of the $j$-th piece of the trajectory, $E_v, E_a, E_c$ are the discrete approximation of the penalty terms respectively, and each piece of the trajectory is discretized into $L$ constraint points. Since $\mathcal{S}$  is a polynomial composed of $\mathbf{C}$ and $\mathbf{T}$, its derivative can be explicitly and easily calculated. 
%Therefore, we mainly deduct the derivative for the penalty terms $E_v,E_a,E_c$.The penalty term and its derivative are always zero if constraints are not violated.
 Thus, we only need to derive the gradient of the point that violates constraints, denoted as
$\partial {E}_{v_{jl}} / \partial \mathbf{C}$,
$\partial {E}_{v_{jl}} / \partial \mathbf{T}$,
$\partial {E}_{a_{jl}} / \partial \mathbf{C}$,
$\partial {E}_{a_{jl}} / \partial \mathbf{T}$,
$\partial {E}_{c_{jl}} / \partial \mathbf{C}$,
$\partial {E}_{c_{jl}} / \partial \mathbf{T}$.
Moreover, we denote  the velocity, acceleration, jerk and  $i$-th vertex ${\hat v}^i$ of the bounding volume at the point as $\mathbf{v}$, $\mathbf{a}$, $\mathbf{j}$ and $\bm{\hat \nu^i}={\hat v}^i_{j}(\frac{l}{L}\delta_j)$ to simplify the formula. Then, we first derive the gradient w.r.t. the coefficient matrix $\mathbf{C}$:
\begin{small}
\begin{align}
\frac{\partial {E}_{v_{jl}}}{\partial \mathbf{C}}
&=3W_v\frac{\delta_j}{L} (||\mathbf{v}||_2^2 - v_{max}^2)^2
\frac{\partial \mathbf{v}^{\rm T} \mathbf{v}}{\partial \mathbf{C}} \nonumber\\
&=3W_v\frac{\delta_j}{L} (||\mathbf{v}||_2^2 - v_{max}^2)^2
\left[
\begin{matrix}
\mathbf{0}_{2(j-1)s \times 3}
\\
2 \beta^{(1)}(\frac{l}{L}\delta_j){\mathbf{v}}^{\rm T} \\
\mathbf{0}_{2(M-j)s \times 3}
\end{matrix}
\right],
\label{equ:Ev/C}\\
\frac{\partial {E}_{a_{jl}}}{\partial \mathbf{C}}
&=3W_a\frac{\delta_j}{L} (||\mathbf{a}||_2^2 - a_{max}^2)^2
\frac{\partial {\mathbf{a}}^{\rm T} \mathbf{a}}{\partial \mathbf{C}} \nonumber\\
&=3W_a\frac{\delta_j}{L} (||\mathbf{a}||_2^2 - a_{max}^2)^2
\left[
\begin{matrix}
\mathbf{0}_{2(j-1)s \times 3} \\
2\beta^{(2)}(\frac{l}{L}\delta_j){\mathbf{a}}^{\rm T} \\
\mathbf{0}_{2(M-j)s \times 3}
\end{matrix}
\right],\label{equ:Ea/C}\\
\frac{\partial {E}_{c_{jl}}}{\partial \mathbf{C}}&=W_c\sum_{i=1}^{K}\sum_{k=1}^{N_j}  \frac{\partial 	{[{(\vec{n}^k_j)}^{\rm T}(\bm{\hat \nu^i}-  \widetilde{p}^k_j)]}^3}{ \partial \mathbf{C}}\frac{\delta_j}{L} \label{equ:Ec/C} \nonumber\\
&=3W_c\frac{\delta_j}{L}\sum_{i=1}^{K}\sum_{k=1}^{N_j} [{(\vec{n}^k_j)}^{{\rm T}}(\bm{\hat \nu^i}-  \widetilde{p}^k_j)]^2
\frac{\partial {(\vec{n}^k_j)}^{{\rm T}}\bm{\hat \nu^i}}{ \partial \mathbf{C}}.
\end{align}
\end{small}Also, the gradient w.r.t. time allocation vector $\mathbf{T}$ is as follows:
\begin{small}
	\begin{align}
	\frac{\partial {E}_{v_{jl}}}{\partial \mathbf{T}}
%	&=W_v\frac{\partial {(||\mathbf{v}||_2^2 - v_{max}^2)}^3 \frac{\delta_j}{L}}{\partial \mathbf{T}} \nonumber\\
	&=W_v \frac{{(||\mathbf{v}||_2^2 - v_{max}^2)}^2}{L}
	\left[\begin{matrix}
	\mathbf{0}_{j-1}\\
	{(||\mathbf{v}||_2^2 - v_{max}^2)}+\frac{6\delta_jl}{L} \mathbf{v}^{\rm T} \mathbf{a}     \\
	\mathbf{0}_{M-j}
	\end{matrix}\right],
	\label{equ:Ev/T}\\
	\frac{\partial {E}_{a_{jl}}}{\partial \mathbf{T}}
%	&=W_a\frac{\partial {(||\mathbf{a}||_2^2 - a_{max}^2)}^3 \frac{\delta_j}{L}}{\partial \mathbf{T}} \nonumber\\
	&=W_a \frac{{(||\mathbf{a}||_2^2 - a_{max}^2)}^2}{L}
	\left[\begin{matrix}
	\mathbf{0}_{j-1}\\
	{(||\mathbf{a}||_2^2 - a_{max}^2)}+\frac{6\delta_jl}{L} \mathbf{a}^{\rm T} \mathbf{j}     \\
	\mathbf{0}_{M-j}
	\end{matrix}\right],
	\label{equ:Ea/T} \\
	\frac{\partial {E}_{c_{jl}}}{\partial \mathbf{T}}
	&=W_c\sum_{i=1}^{K}\sum_{k=1}^{N_j}
	\left[
	\begin{matrix}
	\mathbf{0}_{j-1} \nonumber \\
	\frac{{[{(\vec{n}^k_j)}^{\rm T}(\bm{\hat \nu^i}-\widetilde{p}^k_j)]}^3}{L} \\
	\mathbf{0}_{M-j}
	\end{matrix}
	\right]+\\
	&3W_c\sum_{i=1}^{K}\sum_{k=1}^{N_j}\frac{\delta_j}{L}[{(\vec{n}^k_j)}^{{\rm T}}(\bm{\hat \nu^i}-  \widetilde{p}^k_j)]^2 \frac{\partial {(\vec{n}^k_j)}^{{\rm T}}\bm{\hat \nu^i}}{ \partial \mathbf{T}}.
	\label{equ:Ec/T}
	\end{align}
\end{small}
%\begin{small}
%	\begin{align}
%	&\frac{\partial {(\vec{n}^k_j)}^{{\rm T}}{\hat v}^i_{j}(\frac{l}{L}\delta_j)}{ \partial \mathbf{T}}=
%	\frac{\partial {(\vec{n}^k_j)}^{\rm T}{R_b}_j(\frac{l}{L}\delta_j)q_v^i }{\partial \mathbf{T}}+
%	\left[
%	\begin{matrix}
%	\mathbf{0}_{j-1}\\
%	\frac{l}{L}{(\vec{n}^k_j)}^{\rm T}\mathbf{v}
%	\\
%	\mathbf{0}_{M-j}
%	\end{matrix}
%	\right],
%	\end{align}
%\end{small}
%\begin{small}
%	\begin{align}
%	&\frac{\partial {(\vec{n}^k_j)}^{\rm T}{R_b}_j(\frac{l}{L}\delta_j)q_v^i }{\partial \mathbf{T}}	=
%	\frac{\sum_{\mu \in \{x,y,z\}}{q_v^i}_{\mu} {(\vec{n}^k_j)}^{\rm T} r_{{b\mu}_j}(\frac{l}{L}\delta_j)}{\partial \mathbf{T}} \nonumber\\
%	&=\left[
%	\begin{matrix}
%	\mathbf{0}_{j-1} \\
%	\sum\limits_{\mu \in {x,y,z}} {q_v^i}_{\mu}\frac{l}{L}\frac{||\mathbf{r}_\mu||_2^2
%		{(\vec{n}^k_j)}^{\rm T}B_\mu\mathbf{j}
%		+ {(\vec{n}^k_j)}^{\rm T} \mathbf{r}_\mu  \mathbf{r}_\mu^{\rm T}B_\mu\mathbf{j}
%	}
%	{||\mathbf{r}_\mu||_2^3}
%	\\
%	\mathbf{0}_{M-j}
%	\end{matrix}
%	\right].
%	\end{align}
%\end{small}
%\hspace{-0.15cm}
\begin{figure}[t]
	\centering
	\begin{subfigure}{0.8\linewidth}
		\centering
		\includegraphics[width=1\linewidth]{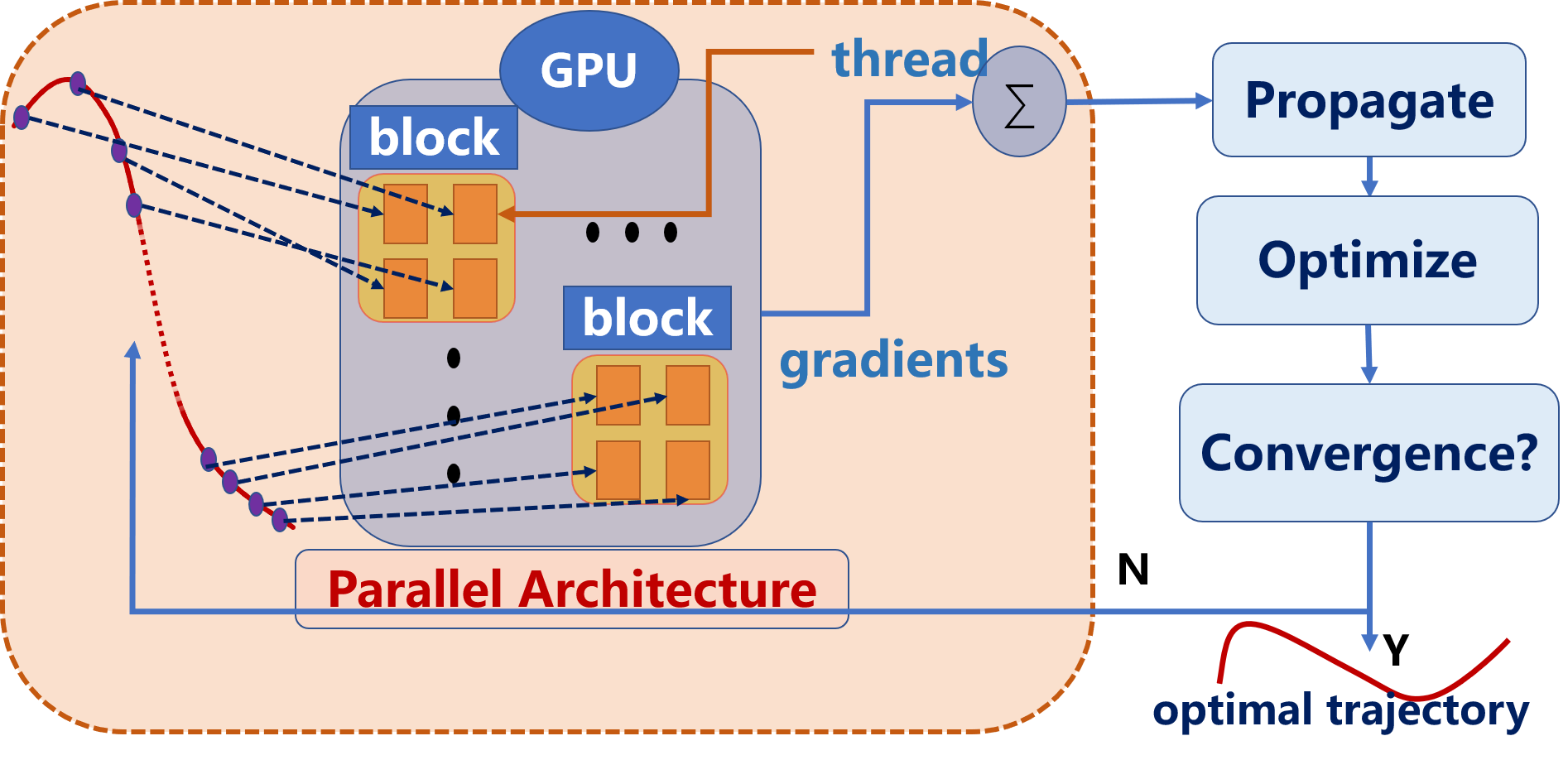}
		\captionsetup{font={small}}
		\caption{The whole optimization process.}
		%\label{pic:urban}
	\end{subfigure}
	\begin{subfigure}{0.8\linewidth}
		\centering
		\includegraphics[width=1\linewidth]{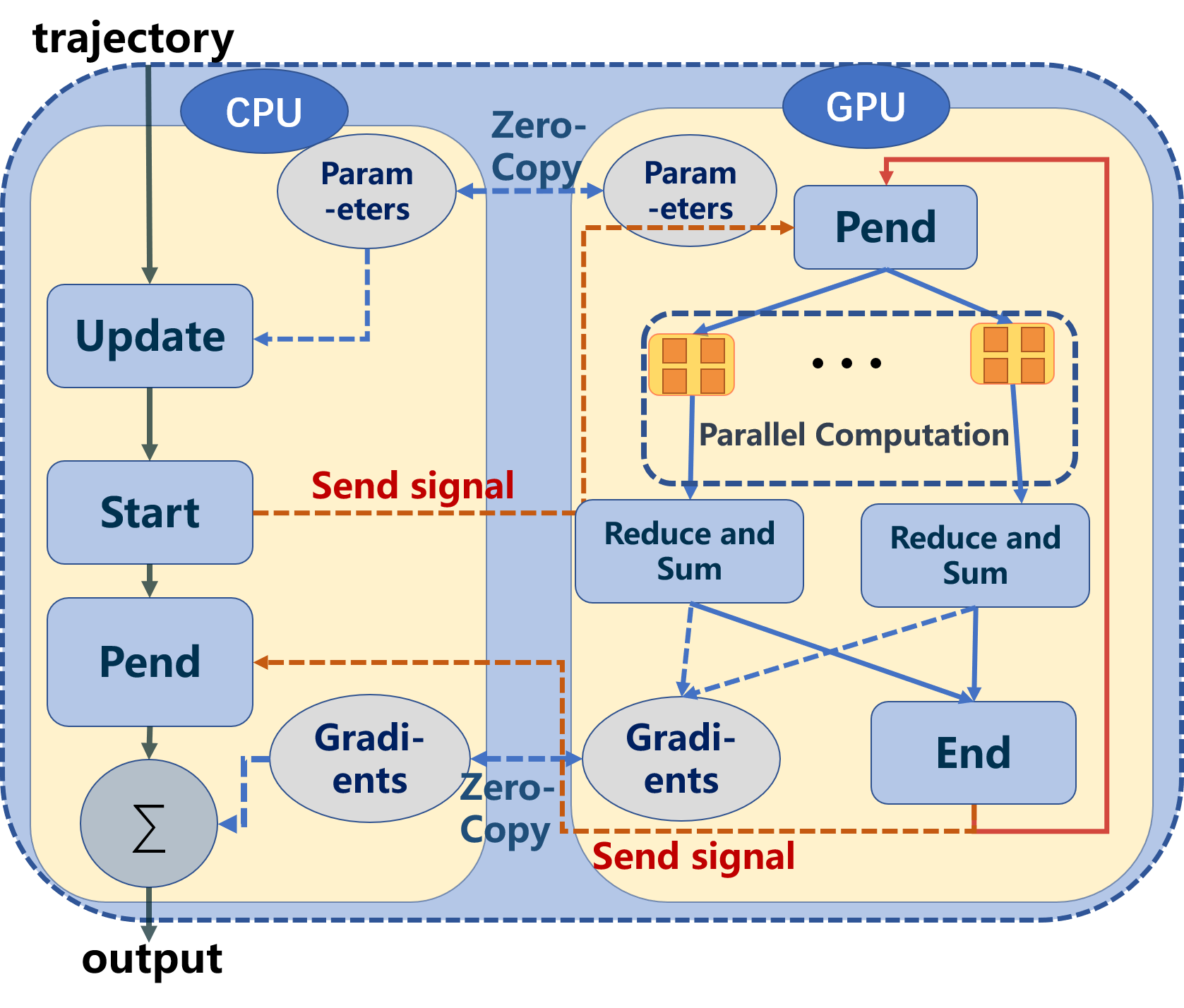}
		\captionsetup{font={small}}
		\caption{The parallel architecture.}
		%\label{pic:top_graph_b}
	\end{subfigure}
	\captionsetup{font={small}}
	\caption{
	An overview of our parallel optimization process.
	}
	\vspace{-1.0cm}
	\label{pic:pararc}
\end{figure}
\begin{figure*}[t]
	\centering
	\begin{subfigure}{0.31\linewidth}
		\includegraphics[width=1\linewidth]{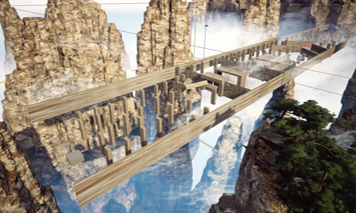}
		\captionsetup{font={small}}
	\end{subfigure}
	\begin{subfigure}{0.31\linewidth}
		\includegraphics[width=1\linewidth]{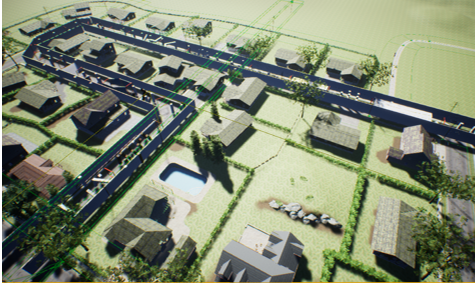}
		\captionsetup{font={small}}
	\end{subfigure}
	\begin{subfigure}{0.31\linewidth}
		\includegraphics[width=1\linewidth]{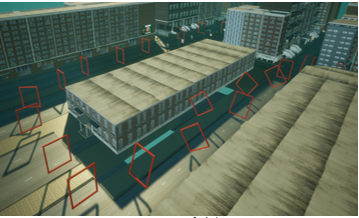}
		\captionsetup{font={small}}
	\end{subfigure}
	\captionsetup{font={small}}
	\caption{
			A panoramic view of our racing environments: Left - Track(a), Middle - Track(b), Right - Track(c).
	}
	\label{pic:racingtracks}
	\vspace{-0.2cm}
\end{figure*}Furthermore, we obtain $\partial H / \partial \mathbf{C}$ and $\partial H /\partial \mathbf{T}$ by substituting 
\eqref{equ:Ev/C}-\eqref{equ:Ec/C} and
\eqref{equ:Ev/T}-\eqref{equ:Ec/T}  to (\ref{equ:H/C}) and (\ref{equ:H/T}) respectively.
Based on the optimality condition proved in \cite{Wang2021GCOPTER}, the minimum control effort piecewise-polynomial trajectory $\mathbf{x}^{*}(t)$ is $d$ times continuously differentiable at each $T_j$ while the interval condition $\overline{p}_j \in \mathbb{R}^{3(\widetilde{d}+1)}$ is fixed, where $2s = \widetilde{d} + d+2$.
We choose $d=4,\widetilde{d}=0$ in our work, which means snap is always
continuous on the whole trajectory and waypoints $\mathbf{q}=(q_1,\cdots,q_{M-1})\in \mathbb{R}^{3\times (M-1)}$ are fixed. Then, $\mathbf{q}$, $\mathbf{C}$, and $\mathbf{T}$ can be related by a non-singular mapping matrix $\mathbf{M}$
defined in ~\cite{Wang2021GCOPTER} that is only associated with  $\mathbf{T}$:
\begin{small}
\begin{align}
& \ \ \ \ \ \ \ \ \ \ \ \ \ \ \ \  \mathbf{M}\mathbf{C} = \mathbf{b} \Rightarrow  \mathbf{C} = {\mathbf{M}}^{-1} \mathbf{b} \label{equ:opticondition}, \nonumber \\
&\mathbf{b}  ={[\overline{p}_0^{\rm T},q_1^{\rm T},\mathbf{0}_{3 \times (d+1)},
	 \cdots,q_{M-1}^{\rm T},\mathbf{0}_{3 \times (d+1)},	\overline{p}_f^{\rm T}
	]}^{\rm T}.
\end{align}
\end{small}Therefore, the trajectory is uniquely determined by $\mathbf{q}$ and $\mathbf{T}$. Then, the problem formulation is transformed to:
\begin{small}
\begin{align}
&{\rm min} \ \hat H(\mathbf{q},\mathbf{T}) = H(\mathbf{C}(\mathbf{q},\mathbf{T}),\mathbf{T}),  \\
&s.t. \ q_j   \in {\mathcal{C}_j \cap \mathcal{C}_{j+1}}, \ \forall j \in \{1,\cdots,M-1 \},  \label{unequ: spacons} \\
&\ \ \ \ \ \delta_j>0 \ \forall j \in \{1,2,\cdots,M\}. \label{unequ: temcons}
\end{align}
\end{small}Moreover, gradients w.r.t. $\mathbf{q}$ and $\mathbf{T}$ are as follows:
\begin{small}
\begin{align}
&\frac{\partial \hat H(\mathbf{q},\mathbf{T})}{\partial \mathbf{q}} = \frac{\partial H}{\partial \mathbf{C}}\frac{\partial \mathbf{C}}{\partial \mathbf{q}},	\\
&\frac{\partial \hat H(\mathbf{q},\mathbf{T})}{\partial \mathbf{T}} = \frac{\partial H}{\partial \mathbf{T}}+\frac{\partial H}{\partial \mathbf{C}}\frac{\partial \mathbf{C}}{\partial \mathbf{T}},
\end{align}
\end{small}where $\partial \mathbf{C} / \partial \mathbf{q}$ and $\partial \mathbf{C} / \partial \mathbf{T}$ can be computed with a linear-complexity way leveraging the result in \cite{Wang2021GCOPTER}. Then, we use a method proposed in \cite{Wang2021GCOPTER} to eliminate the temporal constraints (\ref{unequ: temcons}) and spatial constraints (\ref{unequ: spacons}). The constrained optimization problem
is turned into an unconstrained one that can be solved by quasi-Newton methods efficiently and robustly.
\subsection{High-performance Implementation of $\mathrm{SE}(3)$ Planner}
\label{sec:parallel computing}
In the proposed planning approach, one necessary part is to calculate the gradient for each constraint point.
In each iteration of quasi-Newton method, the gradient calculation for each piece of the trajectory costs constant time $U$ which is related to $L$.
Therefore, the time complexity for calculating gradients of the whole trajectory is $U\cdot M$. Despite the linear complexity, it is a huge cost for large-scale trajectory optimization such as global trajectory generation which is common in drone racing. Fortunately, the gradient calculation for each constraint point on the entire trajectory is independent.
Thus, we calculate the gradients parallelly to solve the problem
more efficiently with GPU acceleration tools.
Then, the computation time for each iteration is reduced to $\frac{U}{L}$.
Moreover, we adopt zero copy which maps host memory to device memory space directly to reduce the explicit data transmission time.
The whole framework is shown in Fig.\ref{pic:pararc}.
The procedures of the parallel architecture are detailed as follows:
\begin{itemize}
	\item [1)]
	Once the optimization problem is set up, we first launch our persistent kernel function which is always running on GPU to reduce the time of repeated activation of the device function.
	\item [2)]
	In each iteration, we update the trajectory parameter information and send a signal from the CPU to the GPU to start the calculation process. The gradient $g_{jl}$ of the $l$-th discretization point in the $j$-th piece of the trajectory is computed in the $l$-th thread of the $j$-th block independently. Moreover,
	We store the result in the local shared memory of each block.
	\item [3)]
	On the GPU, we reduce and sum \cite{harris2007optimizing} the gradients stored in the local memory for each block to
	obtain the gradient $\sum_{l=0}^{L-1} g_{jl}$ for each piece of the trajectory in logarithmic time  complexity.
	\item [4)]
	We detect the parallel computation process continuously on the CPU. After the computation process is finished, the GPU sends the signal to the CPU. Then, we sum the gradient of each piece on the CPU to acquire the final gradients for the entire trajectory.
\end{itemize}
\section{$\mathrm{SE}(3)$ Drone Racing Simulation}
We build simulation environments based on Airsim\cite{shah2017airsim}, a highly integrated and complete simulation framework for multirotor which implements a lightweight physics engine, flight controller, and photorealistic scene. It supports more debugging APIs, sensors and control modes compared with other simulation kits such as Flightmare~\cite{song2020flightmare} and FlightGoggles~\cite{MitFlightGooles}. Furthermore, it is highly convenient for users to create or change the environment in Airsim, which is necessary to build  unique tracks. Significantly, as one of the most popular multirotor simulators, Airsim provides a more detailed document than others, which makes it easier for users to get started with our work. Nonetheless, drone racing utilizing Airsim is not out-of-the-box. For drone racing that focuses on  planning, challenging scenarios, evaluation metrics, a racing progress monitor, a controller and the global map interface are all important to validate a planning module. In Sect.\ref{sec:environment}, we show our tracks for drone racing simulation. In Sect.\ref{sec:droneracing}, we discuss the evaluation criteria and rules of drone racing. In Sect.\ref{sec:mapinteface}, we show common interfaces for users.
\subsection{Simulation Environments}
\label{sec:environment}
Thanks to Unreal Engine (UE) Editor, we are able to build various drone racing scenes, as shown in Fig.\ref{pic:racingtracks}.
%we design extremely dense and complex drone racing tracks containing a set of custom UE actors. These actors detect race-related simulation events that are important to the racing evaluation, such as drone collisions. 
Furthermore, some gates with known poses and sizes are placed in the environment for racing score statistics and direction guides.
\par
Drone racing is a systematic problem that involves lots of aspects such as motion planning, flight control, state estimation, and map construction. However, we hope that our racing is a pure test of motion planning algorithms. The construction of maps is never our concern. As a result, a global map is known for planning. The size and pose of each obstacle are known, which are directly used to set up the global point cloud.
\par
To discriminate performance of different motion planning algorithms, and show the advantage of the $\mathrm{SE}(3)$ planning, there are some narrow gaps in track(a), as shown in Fig.\ref{pic:features}.
Therefore, similar to our planner in Sect.\ref{sec:se3planning}, the shape and attitude of the drone must be taken into consideration to cross the gap, as shown in Fig.\ref{pic:top_graph_b}. For track(b), there are no gaps but extremely dense obstacles, which are difficult for obstacle avoidance. Moreover, to imitate  traditional drone-racing scenes such as \cite{MoonChallenges} and \cite{foehn2020alphapilot}, we build track(c) which only consists of gates for drones to pass.
\subsection{$\mathrm{SE}(3)$ Drone Racing Evalution}
\label{sec:droneracing}
In drone racing, drones are required to pass a series of gates with known poses while avoiding various obstacles, and reach the finish line in the minimum possible time. To collect data for comparison, a monitor program is provided which records the racing time, collision information, the number of passed gates, and the position of the drone in real-time. We present a metric to quantify each racing result. An evaluation score is calculated as follows:
\begin{equation}
S=\left\{
\begin{aligned}
& 100-T_f+4*N_G-P_c 					& if \ collision \\
& 100-T_f+4*N_G  					& else\\
\end{aligned}
\right.
\label{equ:score calculation}
\end{equation}
where $S$ is the score, $T_f$ is the time to finish the whole track, $N_G$ is the number of passed gates, and  $P_c=30$ is the penalty of collision.
\subsection{Common Interfaces}
\label{sec:mapinteface}
We provide common map interfaces such as grid map, Euclidean Sign Distance Field (ESDF) and octree map to free our users from map construction. Moreover, some planners such as \cite{fei2018icra} require free space as the precondition to optimize the trajectory. We also render the optimal path search and the safe space generation interfaces for users. In addition, to track the planned trajectory, a controller which minimizes position, velocity and attitude errors is also offered, so that users only need to focus on the planner to enjoy the drone racing in our simulation environments. 
\begin{figure*}[t]
	\centering
	\includegraphics[width=1\linewidth]{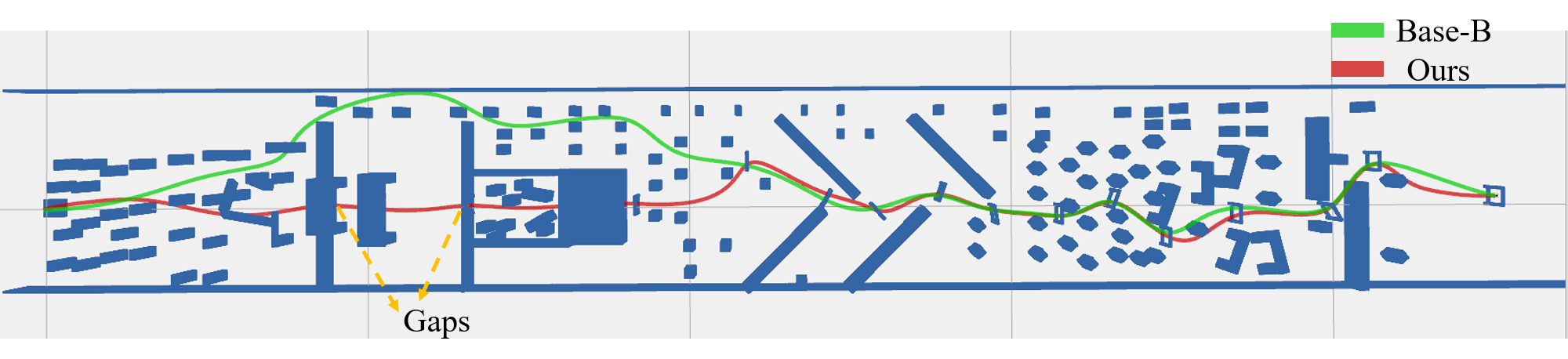}
	\captionsetup{font={small}}
	\caption{
		Comparison of planned  trajectories in track(a).
	}
	\vspace{0cm}
	\label{fig:trackacom}
\end{figure*}
\section{Evaluation}
\label{sec:evaluation}
\subsection{Parallel Architecture}

In this section, we mainly present the efficiency comparison between two different implementations of  $\mathrm{SE}(3)$ planner based on CPU and the proposed parallel architecture.  Then, we show 
the memory overhead of using parallel computing in different tracks.
\par
We conduct the test under different settings on $M$ and $L$, which are defined in Sect.\ref{sec:Problem Formulation} and Sect.\ref{sec:GraCal}, respectively. The trajectory optimization is repeated for more than 200 times in each case. All the tests are conducted in the Linux environment on a computer equipped with an Intel Core i7-10700 CPU and a GeForce RTX 2060 GPU.  
We denote $t_{c}$ and $t_{g}$ as the optimization time of the planner operated on CPU and parallel architecture, respectively. Moreover, we define $r_e = t_c/t_g - 1$ as the efficiency improvement rate.
From Tab.\ref{tab:planner_cmp}, we conclude that the proposed parallel architecture is more efficient than the sequential one in all of these cases, especially for a long trajectory with highly dense collision evaluation. Therefore, the results validate our parallel architecture's
superiority in drone racing where a long global trajectory with 
high-fidelity feasibility detection is often required.
\par
Additionally, in practice, $T_r\cdot N$ cores are used to compute the gradient of  the trajectory which has $N$ pieces. $T_r$ is the number of threads used in each block, which is chosen as 64 in the implementation. Moreover, 4736, 6336 and 3008 CUDA cores are used to generate trajectories  in track(a), (b) and (c), respectively.
\begin{table}[t]
	\centering
	\caption{Comparison of Serial and Parallel Computing.}
	\setlength{\tabcolsep}{1.05mm}
	\renewcommand\arraystretch{1.2}
	{
		\begin{tabular}{|c|c|c|c|c|c|c|c|c|c|}				
			\hline
			Cases & \multicolumn{3}{|c|}{$1\leq M \textless 31$} & \multicolumn{3}{|c|}{$31\leq M \textless 
				65$} & \multicolumn{3}{|c|}{$65\leq M \leq 100$} \\ \hline
			$L$ & $t_{c}(s)$&$t_{g}(s)$ &$r_e$(\%) &$t_{c}(s)$&$t_{g}(s)$&$r_e$(\%)&$t_{c}(s)$&$t_{g}(s)$&$r_e$(\%) \\ \hline
			16 & 1.23 &\bf0.62&\bf98&7.14&\bf3.09&\bf131&14.94&\bf6.43&\bf132 \\ \hline
			32 & 2.45 &\bf0.71&\bf245&12.56&\bf3.44&\bf265&28.94&\bf7.51&\bf285 \\ \hline
			48 & 3.31 &\bf0.73&\bf353&17.22&\bf3.77&\bf357&35.94&\bf7.79&\bf361 \\ \hline
	\end{tabular}}
	\vspace{-0.3cm}
	\label{tab:planner_cmp}	
\end{table}
\begin{table}[t]
	\centering
	\caption{Comparison of Drone Racing Results.}
	\setlength{\tabcolsep}{1.05mm}
	\renewcommand\arraystretch{1.2}
	{
		\begin{tabular}{|c|c|c|c|c|c|c|}				
			\hline
			& \multicolumn{2}{|c|}{Racing Time($s$)} & \multicolumn{2}{|c|}{Score} & \multicolumn{2}{|c|}{Passed Gates}\\ \hline
			Environments &Ours   &Base-B  &Ours   &Base-B&Ours&Base-B\\ \hline
			track(a)     & \textbf{35.49} &38.34   &\textbf{124.5}  &109.65   &\textbf{15}&12 \\ \hline
			track(b)     & \textbf{33.01} &44.21   &\textbf{151.0}  &139.78   &\textbf{21}&21 \\ \hline
			track(c)     & \textbf{18.10} &22.56   &\textbf{157.9}  &153.44   &\textbf{19}&19 \\ \hline
	\end{tabular}}
	\vspace{-1.2cm}
	\label{tab:racingcmp}	
\end{table}
\subsection{Benchmark Comparison}
In this section, we carry out benchmark comparisons of the proposed planner 
against some  cutting-edge methods.
\par
Although the search-based planner~\cite{liu2017searchbased} (Base-A) can generate a  $\mathrm{SE}(3)$ trajectory,
searching for a feasible solution in a large scene consumes unaffordable  time and memory, which is not suitable for long global trajectory generation.
Thus, we compare the racing results of the recent  $\mathbb{R}^3$ 
planner~\cite{S.liuCorridor} (Base-B)  in the proposed tracks with ours.  
Moreover, a common baseline controller is used for the drone to track the trajectory.
As is shown in Tab.~\ref{tab:racingcmp}, our planner performs better in each 
track, revealing the superiority in drone racing. Furthermore,   Base-B can not even generate a trajectory that passes through all gates in track(a) since there are gaps that are narrower than the diameter of the drone. The visualization of track(a) and the comparison is shown in Fig.~\ref{fig:trackacom}.
\par
\begin{figure}[t]
	\centering
	\includegraphics[width=1.0\columnwidth]{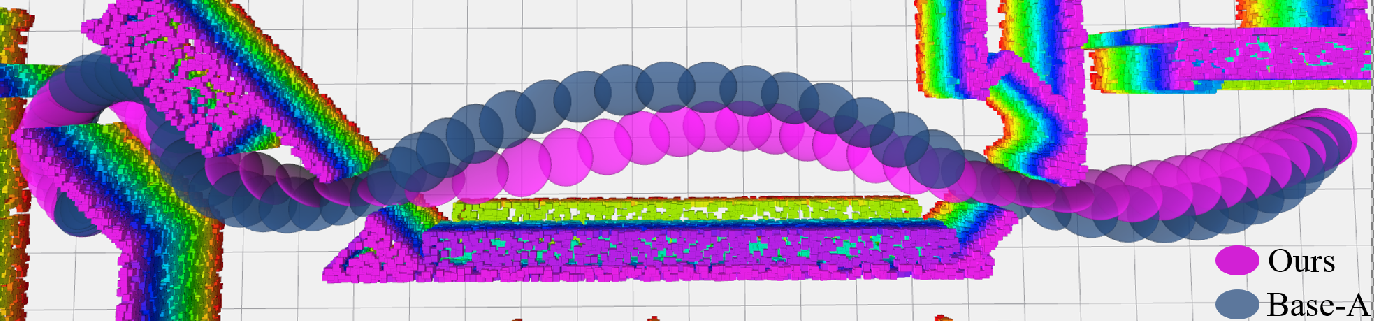}
	\captionsetup{font={small}}
	\caption{Comparison of planned $\mathrm{SE}(3)$ trajectories.}
	\label{fig:benchmarkA}
\end{figure}

In addition, we compare our planner with Base-A in its customized scene. In this comparison,
planners are required to generate a collision-free trajectory from the same initial state to the end state, as shown in  Fig.~\ref{fig:benchmarkA}.  Moreover, we define a loss $J = \int_{0}^{T}{p^{(3)}(t)^{\rm T} p^{(3)}(t)dt+\rho T}$  that focuses on the smoothness
and aggressiveness to  evaluate the performance of planners, where $\rho$ is selected as $1000$.
In this test, the loss $J$ and computation time of Base-A are $12640.0$ and $1617.8ms$ while ours are \textbf{$9640.2$} and \textbf{$163.1ms$}, respectively. 
The result shows that our planner generates a trajectory with higher quality.
Furthermore, our planner also shows a huge time efficiency advantage even if Base-A uses a known prior trajectory.
\par
\begin{figure}[t]
	\centering
	\includegraphics[width=1.0\columnwidth]{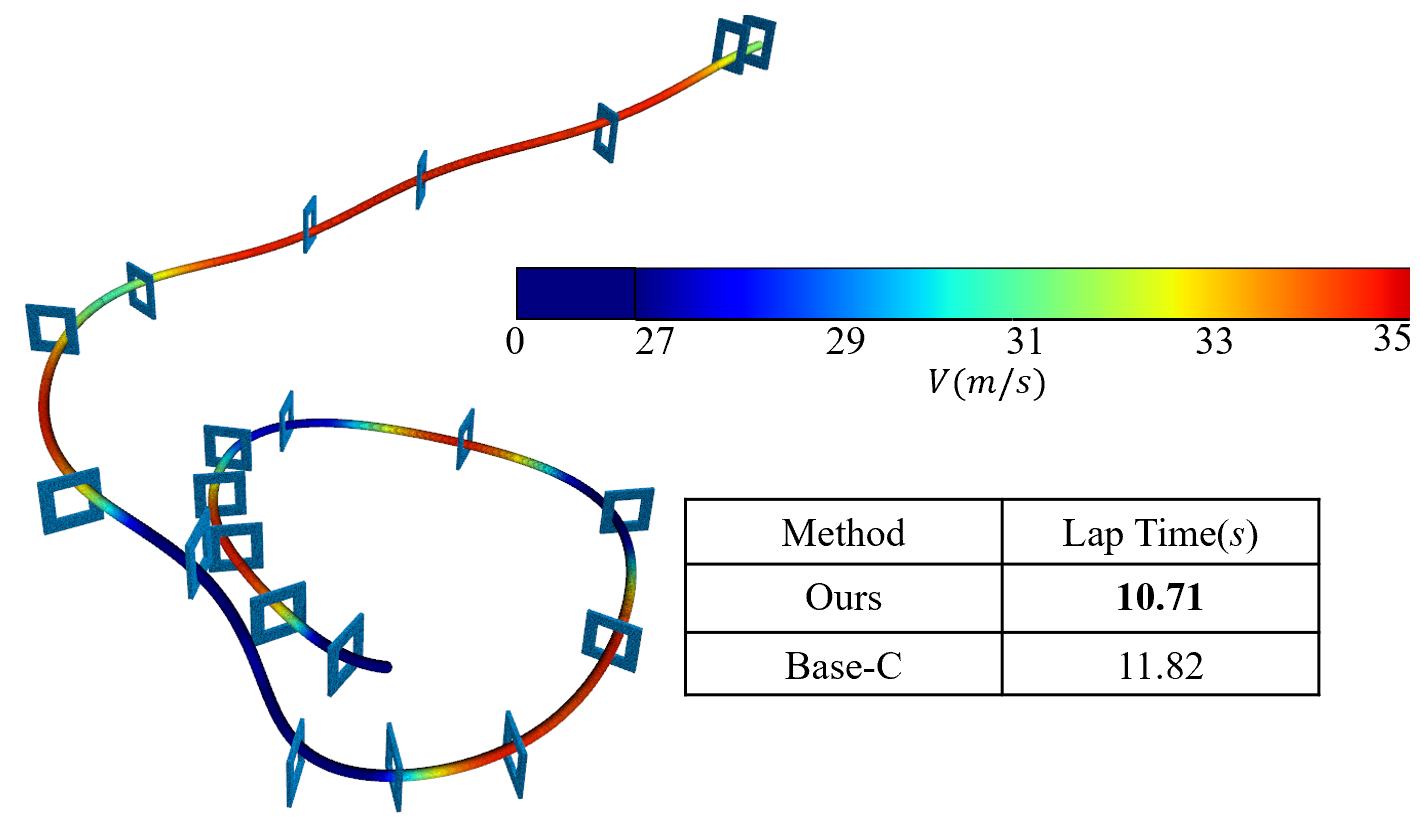}
	\captionsetup{font={small}}
	\caption{The comparison of lap time and the visualization of our planned trajectory.}
	\label{fig:racingcmp}
	\vspace{-2.0cm}
\end{figure}

Furthermore, we also compare our planner with the learning-based method~\cite{song2021autonomous} (Base-C) which specifically focuses on drone racing. 
Since Base-C is not open-sourced now, we compare the racing result in a public track~\cite{madaan2020airsim} with the result in \cite{song2021autonomous}.
Moreover, the common parameters such as maximum velocity and thrust-to-weight ratio are set to the same.
As shown in Fig.\ref{fig:racingcmp}, our planner achieves $9.4\%$ better performance than Base-C. Additionally, in this case, the computation time of our planner is $1.1s$, while Base-C needs to spend a few hours.
\subsection{Real-world Experiment}
\begin{figure}[t]
	\centering
	\includegraphics[width=1.0\columnwidth]{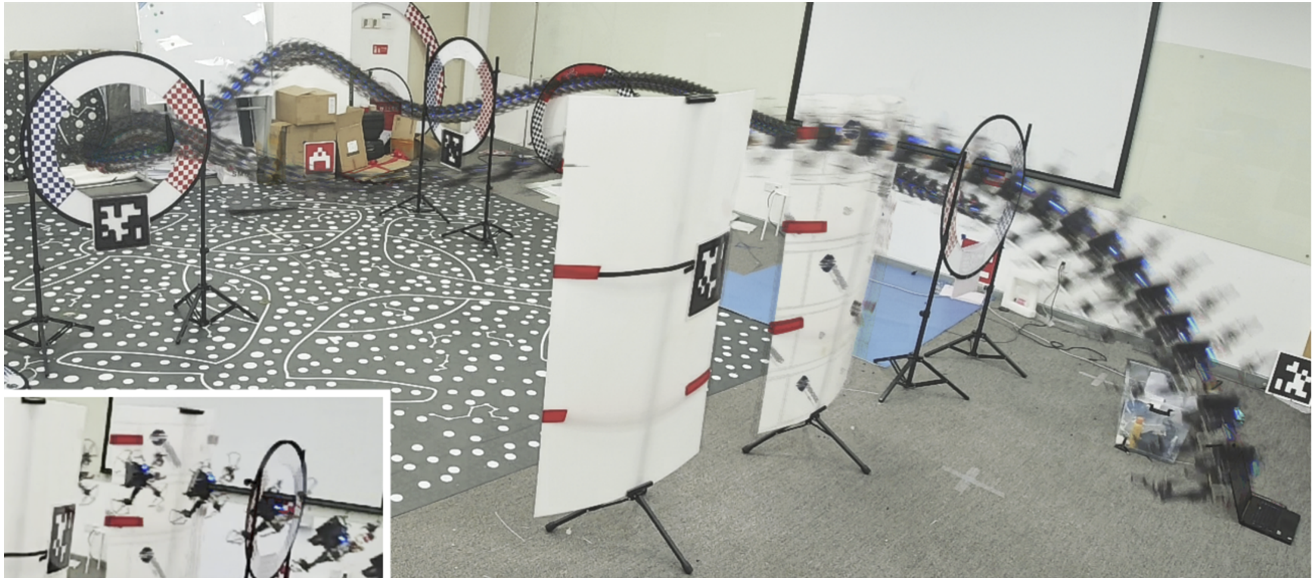}
	\captionsetup{font={small}}
	\caption{The $\mathrm{SE}(3)$ motion in the real-world experiment}
	\label{fig:se3exp}
\end{figure}
We carry out the real-world experiment in a $10m\times 8m$ room with three gates and one gap.
The diameter of the drone is $46cm$, the inner diameter of the gate is $65cm$ , and the width of the gap is $36cm$.
\par
In this experiment, the drone is required to pass through a gap and gates sequentially.
The location of gates is roughly given in advance and then continuously corrected through tags online.
Moreover, we obtain the state of the drone by state-of-the-art VIO system VINS-Fusion~\cite{Qin2018Vins}.
Furthermore, all modules including planner, localization, controller, and perception run onboard in real-time.
Due to localization and control errors, the maximum speed of the drone is set to $4m/s$.
The environment and the real-flight motion are shown in Fig.~\ref{fig:se3exp}.
\section{Conclusion}
\label{sec:conclusion}
In this paper, we present an open-source $\mathrm{SE}(3)$ planning simulator with a strong benchmark baseline, focusing on high-quality and extremely aggressive $\mathrm{SE}(3)$ trajectory generation. We further propose a parallel architecture to speed up the trajectory generation. Our works are user-friendly through their comprehensive interfaces. All works are open-source with the goal of push forward the development of $\mathrm{SE}(3)$ motion planning for multirotor.
\par
In the future, we plan to consider dynamic obstacle avoidance to make the planning baseline applicable to more general scenarios. 
%Last but not least, for more clear comparisons, we will enable the simulation plateform to achieve multiple-agent drone racing, where each drone performs different planning algorhitms simutaneously.
% We enable m
\newlength{\bibitemsep}\setlength{\bibitemsep}{0.0\baselineskip}
\newlength{\bibparskip}\setlength{\bibparskip}{0pt}
\let\oldthebibliography\thebibliography
\renewcommand\thebibliography[1]{%
	\oldthebibliography{#1}%
	\setlength{\parskip}{\bibitemsep}%
	\setlength{\itemsep}{\bibparskip}%
}
%% Use plainnat to work nicely with natbib.
\bibliographystyle{plainnat}
\bibliography{rsshzc}

\begin{thebibliography}{10}
\providecommand{\url}[1]{#1}
\csname url@samestyle\endcsname
\providecommand{\newblock}{\relax}
\providecommand{\bibinfo}[2]{#2}
\providecommand{\BIBentrySTDinterwordspacing}{\spaceskip=0pt\relax}
\providecommand{\BIBentryALTinterwordstretchfactor}{4}
\providecommand{\BIBentryALTinterwordspacing}{\spaceskip=\fontdimen2\font plus
\BIBentryALTinterwordstretchfactor\fontdimen3\font minus
  \fontdimen4\font\relax}
\providecommand{\BIBforeignlanguage}[2]{{%
\expandafter\ifx\csname l@#1\endcsname\relax
\typeout{** WARNING: IEEEtran.bst: No hyphenation pattern has been}%
\typeout{** loaded for the language `#1'. Using the pattern for}%
\typeout{** the default language instead.}%
\else
\language=\csname l@#1\endcsname
\fi
#2}}
\providecommand{\BIBdecl}{\relax}
\BIBdecl

\bibitem{MoonChallenges}
H.~Moon, J.~Martinez-Carranza, T.~Cieslewski, M.~Faessler, D.~Falanga,
  A.~Simovic, D.~Scaramuzza, S.~Li, M.~Ozo, C.~De~Wagter, G.~Croon, S.~Hwang,
  H.~Shim, H.~Kim, M.~Park, T.-C. Au, and s.-j. Kim,
  ``\href{https://link.springer.com/article/10.1007/s11370-018-00271-6}{Challenges
  and implemented technologies used in autonomous drone racing},''
  \emph{Intelligent Service Robotics}, vol.~12, 04 2019.

\bibitem{foehn2020alphapilot}
P.~Foehn, D.~Brescianini, E.~Kaufmann, T.~Cieslewski, M.~Gehrig, M.~Muglikar,
  and D.~Scaramuzza, ``\href{https://arxiv.org/abs/2005.12813}{Alphapilot:
  Autonomous drone racing},'' \emph{arXiv preprint arXiv:2005.12813}, 2020.

\bibitem{LI2020103621}
S.~Li, M.~M. Ozo, C.~De~Wagter, and G.~C. de~Croon,
  ``\href{https://www.sciencedirect.com/science/article/abs/pii/S0921889020304619}{Autonomous
  drone race: A computationally efficient vision-based navigation and control
  strategy},'' \emph{Robotics and Autonomous Systems}, vol. 133, p. 103621,
  2020.

\bibitem{antonini2018blackbird}
A.~Antonini, W.~Guerra, V.~Murali, T.~Sayre-McCord, and S.~Karaman,
  ``\href{https://link.springer.com/chapter/10.1007/978-3-030-33950-0_12}{The
  blackbird dataset: A large-scale dataset for uav perception in aggressive
  flight},'' in \emph{International Symposium on Experimental Robotics}.\hskip
  1em plus 0.5em minus 0.4em\relax Springer, 2018, pp. 130--139.

\bibitem{DelmericoUZH}
J.~{Delmerico}, T.~{Cieslewski}, H.~{Rebecq}, M.~{Faessler}, and
  D.~{Scaramuzza}, ``\href{https://ieeexplore.ieee.org/document/8793887}{Are We
  Ready for Autonomous Drone Racing? The UZH-FPV Drone Racing Dataset},'' in
  \emph{2019 IEEE International Conference on Robotics and Automation (ICRA)},
  2019, pp. 6713--6719.

\bibitem{madaan2020airsim}
R.~Madaan, N.~Gyde, S.~Vemprala, M.~Brown, K.~Nagami, T.~Taubner,
  E.~Cristofalo, D.~Scaramuzza, M.~Schwager, and A.~Kapoor,
  ``\href{https://arxiv.org/abs/2003.05654}{AirSim Drone Racing Lab},'' in
  \emph{NeurIPS 2019 Competition and Demonstration Track}.\hskip 1em plus 0.5em
  minus 0.4em\relax PMLR, 2020, pp. 177--191.

\bibitem{Wang2021GCOPTER}
Z.~Wang, X.~Zhou, C.~Xu, and F.~Gao,
  ``\href{https://arxiv.org/abs/2103.00190}{Geometrically Constrained
  Trajectory Optimization for Multicopters},'' \emph{arXiv preprint
  arXiv:2103.00190}, 2021.

\bibitem{MitFlightGooles}
W.~{Guerra}, E.~{Tal}, V.~{Murali}, G.~{Ryou}, and S.~{Karaman},
  ``\href{https://arxiv.org/abs/1905.11377}{FlightGoggles: Photorealistic
  Sensor Simulation for Perception-driven Robotics using Photogrammetry and
  Virtual Reality},'' in \emph{2019 IEEE/RSJ International Conference on
  Intelligent Robots and Systems (IROS)}, 2019, pp. 6941--6948.

\bibitem{MichaelEuRoc}
M.~Burri, J.~Nikolic, P.~Gohl, T.~Schneider, J.~Rehder, S.~Omari, M.~W.
  Achtelik, and R.~Siegwart,
  ``\href{https://journals.sagepub.com/doi/abs/10.1177/0278364915620033}{The
  EuRoC micro aerial vehicle datasets},'' \emph{The International Journal of
  Robotics Research}, vol.~35, no.~10, pp. 1157--1163, 2016.

\bibitem{Zhu_2018}
A.~Z. Zhu, D.~Thakur, T.~Ozaslan, B.~Pfrommer, V.~Kumar, and K.~Daniilidis,
  ``\href{https://arxiv.org/abs/1801.10202}{The Multivehicle Stereo Event
  Camera Dataset: An Event Camera Dataset for 3D Perception},'' \emph{IEEE
  Robotics and Automation Letters}, vol.~3, no.~3, p. 2032–2039, Jul 2018.

\bibitem{ZurichUrban}
A.~L. Majdik, C.~Till, and D.~Scaramuzza,
  ``\href{http://rpg.ifi.uzh.ch/zurichmavdataset.html}{The Zurich urban micro
  aerial vehicle dataset},'' \emph{The International Journal of Robotics
  Research}, vol.~36, no.~3, pp. 269--273, 2017.

\bibitem{MellingerMiniSnap}
D.~{Mellinger} and V.~{Kumar},
  ``\href{https://ieeexplore.ieee.org/document/5980409}{Minimum snap trajectory
  generation and control for quadrotors},'' in \emph{2011 IEEE International
  Conference on Robotics and Automation}, 2011, pp. 2520--2525.

\bibitem{MichielFlatSystems}
M.~van Nieuwstadt and R.~M. Murray,
  ``\href{https://www.sciencedirect.com/science/article/pii/S1474667017580167}{Real
  Time Trajectory Generation for Differentially Flat Systems},'' \emph{IFAC
  Proceedings Volumes}, vol.~29, no.~1, pp. 2301--2306, 1996, 13th World
  Congress of IFAC, 1996, San Francisco USA, 30 June - 5 July.

\bibitem{fei2018icra}
F.~Gao, W.~Wu, Y.~Lin, and S.~Shen,
  ``\href{https://ieeexplore.ieee.org/document/8462878}{Online safe trajectory
  generation for quadrotors using fast marching method and bernstein basis
  polynomial},'' in \emph{2018 IEEE International Conference on Robotics and
  Automation (ICRA)}.\hskip 1em plus 0.5em minus 0.4em\relax IEEE, 2018, pp.
  344--351.

\bibitem{zhou2019robust}
B.~Zhou, F.~Gao, L.~Wang, C.~Liu, and S.~Shen,
  ``\href{https://ieeexplore.ieee.org/document/8758904}{Robust and efficient
  quadrotor trajectory generation for fast autonomous flight},'' \emph{IEEE
  Robotics and Automation Letters}, vol.~4, no.~4, pp. 3529--3536, 2019.

\bibitem{zhou2020egoplanner}
X.~{Zhou}, Z.~{Wang}, H.~{Ye}, C.~{Xu}, and F.~{Gao},
  ``\href{https://ieeexplore.ieee.org/abstract/document/9309347}{EGO-Planner:
  An ESDF-Free Gradient-Based Local Planner for Quadrotors},'' \emph{IEEE
  Robotics and Automation Letters}, vol.~6, no.~2, pp. 478--485, 2021.

\bibitem{Falanga}
D.~{Falanga}, E.~{Mueggler}, M.~{Faessler}, and D.~{Scaramuzza},
  ``\href{https://ieeexplore.ieee.org/document/7989679}{Aggressive quadrotor
  flight through narrow gaps with onboard sensing and computing using active
  vision},'' in \emph{2017 IEEE International Conference on Robotics and
  Automation (ICRA)}, 2017, pp. 5774--5781.

\bibitem{Loianno}
G.~{Loianno}, C.~{Brunner}, G.~{McGrath}, and V.~{Kumar},
  ``\href{https://ieeexplore.ieee.org/document/7762111}{Estimation, Control,
  and Planning for Aggressive Flight With a Small Quadrotor With a Single
  Camera and IMU},'' \emph{IEEE Robotics and Automation Letters}, vol.~2,
  no.~2, pp. 404--411, 2017.

\bibitem{Hirata}
T.~{Hirata} and M.~{Kumon},
  ``\href{https://ieeexplore.ieee.org/document/6911574}{Optimal path planning
  method with attitude constraints for quadrotor helicopters},'' in
  \emph{Proceedings of the 2014 International Conference on Advanced
  Mechatronic Systems}, 2014, pp. 377--381.

\bibitem{liu2017searchbased}
S.~Liu, K.~Mohta, N.~Atanasov, and V.~Kumar,
  ``\href{https://arxiv.org/abs/1710.02748}{Search-based motion planning for
  aggressive flight in se (3)},'' \emph{IEEE Robotics and Automation Letters},
  vol.~3, no.~3, pp. 2439--2446, 2018.

\bibitem{S.liuCorridor}
S.~{Liu}, M.~{Watterson}, K.~{Mohta}, K.~{Sun}, S.~{Bhattacharya}, C.~J.
  {Taylor}, and V.~{Kumar},
  ``\href{https://ieeexplore.ieee.org/document/7839930}{Planning Dynamically
  Feasible Trajectories for Quadrotors Using Safe Flight Corridors in 3-D
  Complex Environments},'' \emph{IEEE Robotics and Automation Letters}, vol.~2,
  no.~3, pp. 1688--1695, 2017.

\bibitem{harris2007optimizing}
M.~Harris \emph{et~al.},
  ``\href{https://developer.download.nvidia.com/assets/cuda/files/reduction.pdf}{Optimizing
  parallel reduction in CUDA},'' \emph{Nvidia developer technology}, vol.~2,
  no.~4, pp. 1--39, 2007.

\bibitem{shah2017airsim}
S.~Shah, D.~Dey, C.~Lovett, and A.~Kapoor,
  ``\href{https://arxiv.org/abs/1705.05065}{Airsim: High-fidelity visual and
  physical simulation for autonomous vehicles},'' in \emph{Field and service
  robotics}.\hskip 1em plus 0.5em minus 0.4em\relax Springer, 2018, pp.
  621--635.

\bibitem{song2020flightmare}
Y.~Song, S.~Naji, E.~Kaufmann, A.~Loquercio, and D.~Scaramuzza, ``Flightmare: A
  flexible quadrotor simulator,'' in \emph{Conference on Robot Learning}, 2020.

\bibitem{song2021autonomous}
Y.~Song, M.~Steinweg, E.~Kaufmann, and D.~Scaramuzza, ``Autonomous drone racing
  with deep reinforcement learning,'' \emph{arXiv preprint arXiv:2103.08624},
  2021.

\bibitem{Qin2018Vins}
T.~Qin, P.~Li, and S.~Shen, ``Vins-mono: A robust and versatile monocular
  visual-inertial state estimator,'' \emph{IEEE Transactions on Robotics},
  vol.~34, no.~4, pp. 1004--1020, 2018.

\end{thebibliography}
\end{document}